\documentclass[]{style/ceurart}
\usepackage{listings}
\usepackage{subfig}
\usepackage{graphicx}
\usepackage{array}
\begin{document}

\copyrightyear{2024}
\copyrightclause{Copyright for this paper by its authors.
  Use permitted under Creative Commons License Attribution 4.0
  International (CC BY 4.0).}

\conference{CLEF 2024: Conference and Labs of the Evaluation Forum, September 9-12, 2024, Grenoble, France}

\title{Machine Learning for ALSFRS-R Score Prediction: Making Sense of the Sensor Data}

\author[1]{Ritesh Mehta}[
email=rmehta307@gatech.edu,
]
\cormark[1]
\author[1]{Aleksandar Pramov}[
email=apramov3@gatech.edu,
]
\author[1]{Shashank Verma}[
email=sverma342@gatech.edu,
]

\address[1]{Georgia Institute of Technology, North Ave NW, Atlanta, GA 30332}
\cortext[1]{All authors contributed equally.}

% TODO MENTIONS:
% - AP: Need to formulate hypothesis at end of introduction - e.g. with the small dataset the challenges will be with the CV setup, that panel models will be useful

\begin{abstract}
Amyotrophic Lateral Sclerosis (ALS) is characterized as a rapidly progressive neurodegenerative disease that presents individuals with limited treatment options in the realm of medical interventions and therapies. The disease showcases a diverse range of onset patterns and progression trajectories, emphasizing the critical importance of early detection of functional decline to enable tailored care strategies and timely therapeutic interventions.
The present investigation, spearheaded by the iDPP@CLEF 2024 challenge, focuses on utilizing sensor-derived data obtained through an app. This data is used to construct various machine learning models specifically designed to forecast the advancement of the ALS Functional Rating Scale-Revised (ALSFRS-R) score, leveraging the dataset provided by the organizers.
In our analysis, multiple predictive models were evaluated to determine their efficacy in handling ALS sensor data. 
The temporal aspect of the sensor data was compressed and amalgamated using statistical methods, thereby augmenting the interpretability and applicability of the gathered information for predictive modeling objectives. 
The models that demonstrated optimal performance were a naive baseline and ElasticNet regression. 
The naive model achieved a Mean Absolute Error (MAE) of 0.20 and a Root Mean Square Error (RMSE) of 0.49, slightly outperforming the ElasticNet model, which recorded an MAE of 0.22 and an RMSE of 0.50.
Our comparative analysis suggests that while the naive approach yielded marginally better predictive accuracy, the ElasticNet model provides a robust framework for understanding feature contributions.
\end{abstract}


\maketitle

\section{Introduction}

Amyotrophic Lateral Sclerosis (ALS), also known as Lou Gehrig's disease, is a progressive neurodegenerative disorder that affects motor neurons in the brain and spinal cord. The ALS Functional Rating Scale-Revised (ALSFRS-R), widely utilized in clinical and research settings, stands as a critical metric employed by healthcare professionals to evaluate the functional state of ALS patients. The precise forecasting of ALSFRS-R scores is essential for assessing disease progression and the effectiveness of therapeutic measures. Recent developments in sensor technology have opened up new avenues for continuous, non-invasive monitoring of ALS symptoms. The fusion of sensor data with predictive modeling presents the potential for more accurate and timely forecasts of disease progression, significantly benefiting patient care and treatment management.

The iDPP@CLEF 2024 competition is an initiative aimed at leveraging sensor data \cite{BiroloBosoniEtAl2024a} \cite{BiroloBosoniEtAl2024b} and machine learning techniques to predict ALS progression. Task 1 involves predicting the ALSFRS-R scores assigned by medical professionals using sensor data collected via a dedicated app. Task 2 focuses on predicting self-assessment scores recorded frequently by patients. These tasks aim to enhance the accuracy and timeliness of ALS symptom monitoring and forecasting, providing valuable insights for patient care and treatment management. We hypothesized that the relatively small dataset, coupled with a high number of features, will pose significant challenges in our modeling efforts, necessitating the use of strong regularization techniques to avoid overfitting.

To address the prediction of ALSFRS-R scores, we implemented several techniques. 
We started with a naive model to form a baseline and establish a reference for comparison.  The naive model simply carries the last observed value forward. We then explored various Machine Learning algorithms for regression, as well as a Long Short-Term Memory (LSTM) neural network, to model the temporal dependencies in the sequential sensor data. Performance was evaluated via the Root Mean Squared Error (RMSE) and Mean Absolute Error (MAE) metrics identified by the challenge organisers. The main outcome of our analysis confirmed the initial hypothesis, revealing that the small dataset size indeed complicated the modeling process, but effective regularization strategies, particularly through ElasticNet, were helpful. 

The paper is structured as follows: Section 2 reviews related work, providing context and background on ALSFRS-R score prediction and the integration of sensor data with machine learning techniques. Section 3 details the methodology, including data processing steps and the modeling approaches we employed. Section 4 presents the results of our experiments, comparing the performance of different models and discusses our findings. Section 5 outlines future work directions, and Section 6 concludes the study.

\section{Related Work}{\label{sec:relatedwork}}

% The prediction of ALSFRS-R (Amyotrophic Lateral Sclerosis Functional Rating Scale - Revised) scores is a critical area of research in the management and treatment of ALS. 
% Accurate prediction models can significantly enhance patient care by enabling timely interventions and personalized treatment plans. 
In recent years, the integration of sensor data and machine learning (ML) techniques \cite{Gupta2023AthomeWA} has shown promising results in improving the accuracy and reliability of ALSFRS-R score predictions. \cite{Vieira2022AMB} developed machine learning models to objectively measure ALS disease severity using voice samples and accelerometer data, while \cite{pancotti2022deep} further focuses specifically on on deep learning methods to predict ALS disease progression. Outside of the ALS prediction field, \cite{Choi2015DoctorAP} demonstrated the potential of temporal models in the healthcare domain by integrating EHR data; 
\cite{Vidal2023ComparativePA} applied various ML models to sensor data from accelerometers attached to dairy cattle for disease prediction and \cite{Smedley2018LongitudinalPI} demonstrates how temporal patterns from clinical and imaging data can be used to predict residual survival for cancer patients.

A more technical field of the literature deals specifically with the longitudinal aspect of the data and its effect on ML and DL methods. The recent wider application of ML methods in biomedical data has necessitated adapting traditional models to handle repeated measurements over time. \cite{pinheiro2006mixed}. \cite{hajjem2014mixed} extends Random Forests to handle fixed and random effects. \cite{cascarano2023machine, hu2023predictions}  give a more general overview of existing methods, while \cite{wortwein2023neural} focuses specifically on a neural network adaptation that can handle fixed and random effects.

 \section{Methodology}\label{sec:methodology}

 \subsection{Data Description and Preprocessing}
 The raw features dataset provided for the competition consists of two parts: static data and sensor data. The former contains patient-specific baseline (constant) data, such as sex, age etc. The latter contains 90 time-series of patient-specific sensor data, collected over an average of nine months through a dedicated app developed by the BRAINTEASER project, using a fitness smartwatch in the context of clinical trials. 
For some patients, the clinical data starts before the app data, and some patients whose app data goes beyond the last observed clinical data. This required some synchronization between the clinical and the app data, in order to avoid look-ahead bias. Let $t_{1}^{s}$ be the first sensor time point in the raw data, $t_{1}^{c}$ be the first clinical time point in the raw data, as defined by the days from diagnosis for a given patient. Overall, we consider the time-overlap between clinical and sensor data. 
Some special cases are handled as follows:
\begin{itemize}
\item If $t_{1}^{s}$ > $t_{1}^{c}$, then we discard clinical data prior to one step back from $t_{1}^{s}$. For example, consider a patient that has clinical observations at days 690, 780, 873, and sensor data from 800 onwards. We discard the clinical observation 690, but keep the rest. As we will see later, the reason for this is that we will use the previous clinical score as a feature and hence the first value to be predicted would be the target at day 873. That first value will be predicted using the sensor data between 800 and 872, as well as previous clinical visit from the 780th day since diagnosis.
\item If $t_{1}^{s} \leq t_{1}^{c}$, then we use $t_{1}^{s}$ as the starting point for the sensor data and we do not discard any clinical data. 
\item Sensor data after the last clinical observation is discarded.
\item If the clinical data ends after the sensor data, we consider that point only if it is at maximum of 60 days from the last observed sensor data point. This is done so that we do not use too distant sensor data as a feature.
\item Some patients have only one clinical observation. As we will use the previous clinical question responses in the set of features for the models, those patients will be discarded due to the missing previous response value.
\end{itemize}
Some patients had a few missing static data observations as well. Those were imputed by a simple median over all other patients for the respective feature.

\begin{figure}[!h]%
    \centering
    \includegraphics[scale=0.4]{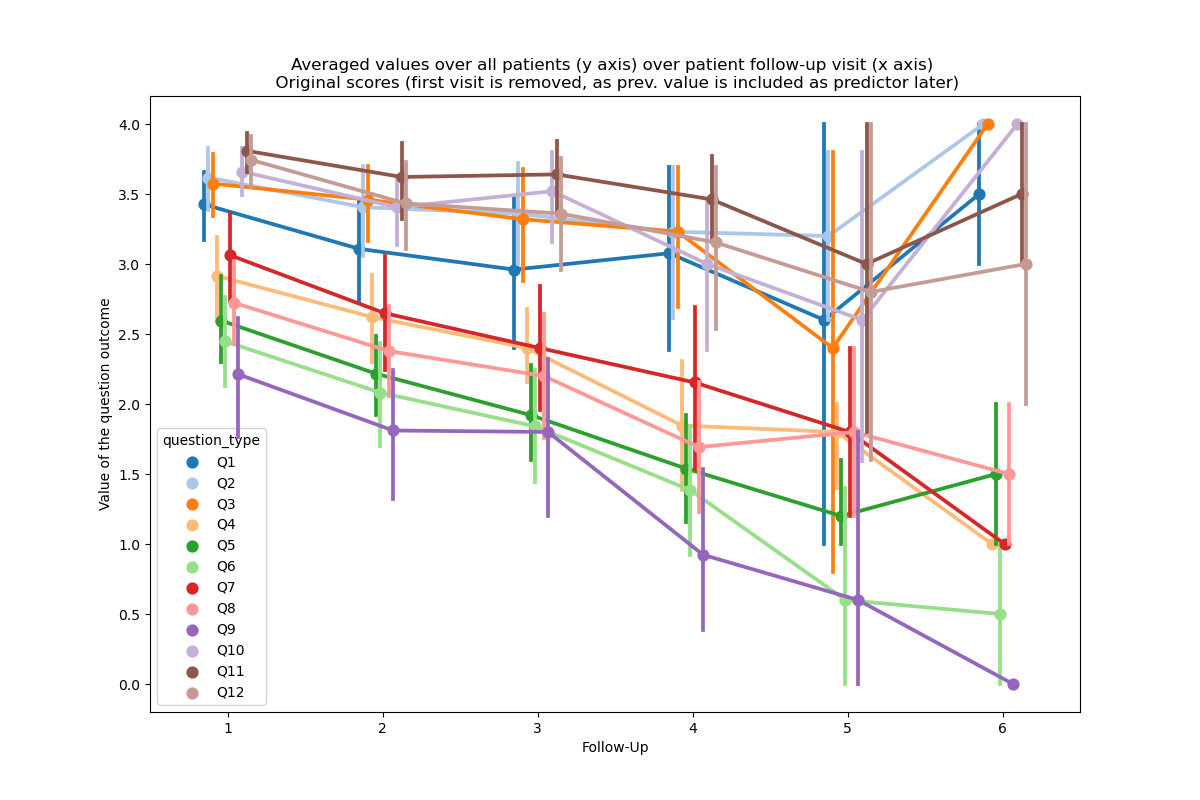}
    \caption{Averaged question response values for each follow-up time point. The x-axis shows the ordered visits by patient, whereas the y axis shows the average score. The vertical bars indicate a 95\% CI around the estimated mean. As we have fewer and fewer patients for longer follow-ups, the CI widen. There is some degree of variability with respect to the intercept among the questions, as well as the slopes. The beginning periods exhibit slow deterioration which already suggests that using the previous value as predictor for the next score will be useful. The points are jittered for visualization purposes.}%
    \label{fig:follow_up_values}%
\end{figure}

 The target variable (i.e., questionnaire responses) provides highly informative insights when visualized and analyzed over time, as shown in Figure \ref{fig:follow_up_values}. The visualization is done on the actual dataset we will use, i.e. after the pre-processing steps described above.
 
 The curves reveal an initial period of slow degradation followed by a more rapid decline for many questions. We also notice that the scores change pretty slowly for most patients over the course of ~100 days (the average number of days between consecutive visits to the clinician). Additionally, only a few patients manage to recover their scores to better levels. That observation has to be cautioned by the observation that the amount of patients with a longer follow-up decreases strongly over time. While at the beginning there are $n=51$ observed patients (i.e. the full sample), this number drops to $n=37$ for the second follow-up and decreases further to just $n=5$ and $n=2$ for the fifth and sixth follow-up respectively. This is also evident by the higher CI (Confidence Interval) bars on Figure \ref{fig:follow_up_values}.

 A key insight here, as a result of above mentioned observations, is that the previous value of the score could be a useful engineered feature, which will also dictate the data pre-processing steps. Additionally, the heterogeneity in question types supports two modeling approaches: (1) treating the data as panel data with ``question type" as a grouping variable for random effects, and (2) modeling each question separately.

\subsection{Features generation}
\label{sec:features_generation}

For our modeling, we considered three groups of features: \emph{target-based}, \emph{static}, and \emph{sensor-data}. While we used the static data \emph{as-is}, we originally tried to build a baseline model based only on the target data and treat it as an autoregressive problem with some additional engineered simple transformations of the target dataset, e.g., the difference in days between two diagnosis, the first value of each patient, the previous value, one-hot encoding of the question type, the follow-up time-index (i.e. the x-axis in Figure \ref{fig:follow_up_values}). Based on this initial analysis, we quickly realized our initial hypothesis for the importance of the previous target value, stated in previous section, is valid, and we kept it as a feature for any subsequent modeling.  

A central challenge in this study was the handling of the \emph{sensor data} features. Those are observed daily (with some gaps), compared to the target (and related engineered variables such as the previous value), which is observed once every three months. To address this mismatch in the frequencies, we followed three different approaches: 
\begin{description}
\item[Approach 1:] By simply taking the median over a window of each sensor data column, where the window was defined to be between each two consecutive clinical visits - $[t-1, t)$.
\item[Approach 2:] By generating a vast array of features derived from a window in each sensor data column, where the window is between each two consecutive clinical visits - $[t-1, t)$, inspired by a feature-based time series analysis approach \cite{fulcher2018feature}.
\item[Approach 3:] By using LSTM (Long Short Term Memory) cells for the sensor data and letting the network define the relevant transformations of the input. 
\end{description}
Note that Approach 1 is a special and simple case of Approach 2. Here, we effectively express each sensor time window $[t-1, t)$, with just one number and a set of numbers respectively. Doing this across all the windows of all patients builds the sensor data feature set.
Approach 3 on the other hand does not have pre-defined (set of) transformation(s) on the input sensor data and we let the neural network itself pick up the relevant transformation of the input. Its details are provided in section 3.5 where the LSTM model is described. \\
For Approach 2, we employed the \texttt{tsfresh} python library for systematic feature engineering from time-series and other sequential data \cite{christ2018time}. Simple examples of those can be e.g. the maxima/minima/median/mean etc. over each sensor data window per each patient between two clinical visits. A full outline of the feature extraction procedure and the considered extracted features can be found on the documentation website of the package. 
Note that, when one considers different parameter setups of the various extracted features, one would end up with multiple extracted features per time series. We have around 100 sensor time series and many of the extracted features by \texttt{tsfresh} had problems like high amount of missing values, low variability etc. and so were removed from the set of feature candidates. On each of those extracted features we performed filtering, by calculating \texttt{tsfresh} Spearmann correlation coefficient between the feature and the (one-step-ahead) clinical values per each question, over all patients. The hypothesis tests for significance are adjusted following the procedure in \cite{DBLP:journals/corr/ChristKF16}.  For the final feature set based on the app data, we experimented with either a) keeping all the features that were deemed significant per question or b) by fixing the number of the top k (e.g. 10) in terms of lowest p-value to keep in the sensor feature set.
As the final modeling was done for each question separately, we settled on the former choice.

\subsection{Data augmentation}
\label{sec:data_augmentation}
\begin{figure}%
    \centering
    \subfloat[\centering In agreement]{{\includegraphics[width=4cm]{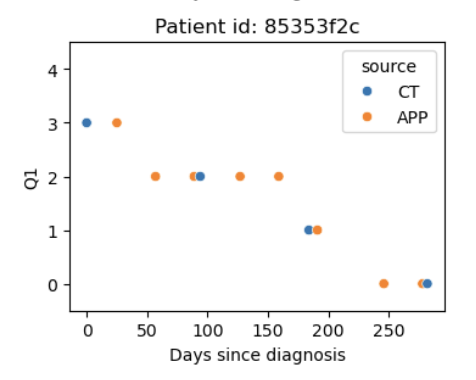} }}%
    \qquad
    \subfloat[\centering Makes patient usable]{{\includegraphics[width=4cm]{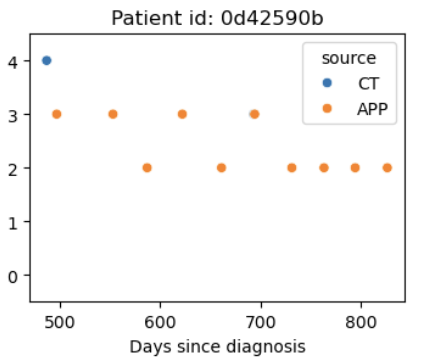} }}%
    \qquad
    \subfloat[\centering Not in agreement]{{\includegraphics[width=4cm]{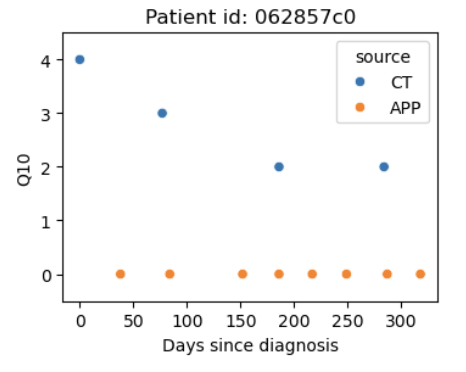} }}%
    \caption{Augmenting Task1's CT scores data with Task2's APP scores. a) An example of patient\_id, question\_id pair where APP scores are in agreement with CT scores b) An example of patient\_id, question\_id pair where adding APP scores allows the usability of this datapoint as it had only 1 CT score data c) An example of patient\_id, question\_id pair where the APP scores aren't in agreement with CT scores.}%
    \label{fig:example}%
\end{figure}

As already mentioned, one of the major challenges in modeling for Task1 was the lack of sufficient datapoints.
% We were given a total of 52 patients with an average of <4 visits per patient resulting in 129 tuples of (score, delta\_days, future\_score)
 Given the high amount of features, this made the models more prone to overfitting . Fitting complex deep models would exacerbate this problem further. To circumvent this issue, we decided to augment the Task1 data with that of Task2. (\emph{Note that doing this reduces the average number of days between any two consecutive visits.}) This gave us ALSFRS-R scores progression as shows in Figure \ref{fig:example}. There are a couple of advantages of doing this:
\begin{itemize}
    \item adding more datapoints for training (more labeled data per patient, question tuple) as shown in Figure\ref{fig:example}(a)
    \item allowing certain patient's data to be not discarded entirely (due to the patient having only 1 visit to the clinician i.e. CT data) in an already short list of patients as shown in Figure\ref{fig:example}(b)
\end{itemize}
However, we cannot simply add this data unilaterally as there are a couple of concerns:
\begin{itemize}
    \item adding the data from patient's self assigned scores results in consecutive scores being too close to each other.
    \item potential disagreement between the scores assigned by the clinician and that assessed by the patient resulting in cases like Figure\ref{fig:example}(c).
\end{itemize}
The solve the first problem, we simply restricted training and prediction to target scores roughly 100 days ahead, instead of using the very next available score.
The second problem requires statistical analysis on a per patient, question pair to decide whether or not to augment this prediction task with patient's self assessment scores. We used the chi-squared test \cite{Pearson1900} to test for the Null the data from the two time series are sampled from the same distribution. Of the 610 patient, question pairs where this analysis was possible, we failed to reject the Null 559 times and thus their data was merged. For the subsequent analysis, we used either only Task1 data or the combined Task1+Task2 data, and we indicate which data set was used where applicable.

\subsection{Modeling}
Unless explicitly specified, all modeling efforts described are for Task1 i.e. predicting clinician's assigned scores. Our modeling setup used RMSE as the scoring function at all stages.
% \begin{enumerate}
%     \item CV setup (Alex)
%     \item ML 'one-to-one' modeling we have applied on tabular data (Shashank)
%     \item NN LSTM 'many-to-one' modeling as it requires different handling  (Ritesh)
% \end{enumerate}

\subsubsection{CV setup}
The small dataset (129 datapoints per question only) posed several challenges with regards to choosing the appropriate cross-validation procedure for the hyperparameter tuning for our models. One key aspect is that the true (unseen) test set contains patients that are not present in the training set, rather than containing unseen data of the same patients. 
Hence, to assess the ability of our models to generalize well, we used a nested k-fold cross validation strategy \cite{cawley2010over, kuhn2019feature}. This consists of two loops - an inner, and an outer one.

In each inner loop, we set aside a test set of 10\% containing complete data of the patients (i.e. all observations). For the remaining data, we perform a further k-fold cross-validation, whereby we also adapted it to contain a complete set of observations of each patient in both the training and the validation sets. That ensured that there is no information leakage between different times of the patient's data, as the true out of sample test (for the final submission) also contains data on unseen patients.

The outer loop then repeats the same procedure, but on another test set which covers another 10\% of the patients. In total, we have 10 iterations in the outer loop, for each of which the RMSE is computed to allow for a model choice. The best hyperparameters overall are then chosen and all the data was fit using those prior the final submission.

\begin{figure}%
    \centering
     \includegraphics[width=0.7\textwidth, height=0.35\textheight]{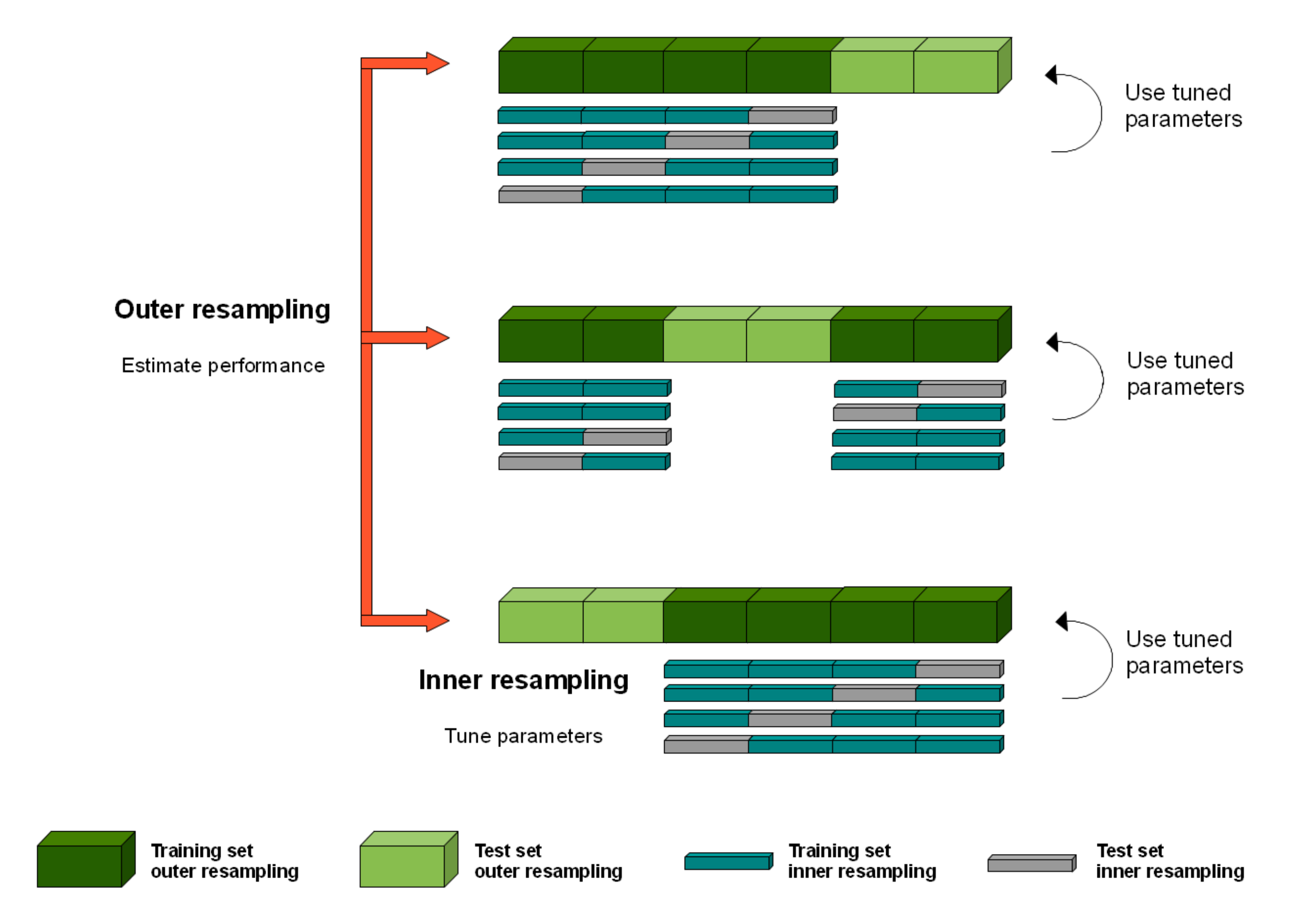}
    \caption{Cross-validation procedure for the hyperparameter tuning and final model fit in our work (Source: \cite{lyashenko_jha_2024}). The original source illustrates the procedure without taking into account a grouped structure as in our dataset. Hence, in our work we adapted the procedure so that each patient's complete data is contained in either the inner-train/test subsets, or the outer test-set.}%
    \label{fig:CV}%
\end{figure}

Figure \ref{fig:CV} illustrates the process for non-grouped data as described by \cite{lyashenko_jha_2024}. The term ``test inner-resampling" corresponds to what we call validation set. Overall, the additional ``grouped" adaptation that we implemented to prevent information leakage, ensured that for any outer-loop/inner-loop combination, all the data for a given patient is contained in either the (outer-loop) test or the (inner-loop) train + validation set. 

\subsubsection{Traditional ML models}
We experimented with a wide range of modeling techniques for predicting ALSFRS-R scores. This included traditional ML models as well as deep learning models, features described in Approach 1 as well as Approach 2 mentioned in section 3.2, with and without augmenting the training data from Task1 with that of Task2 as described in section 3.3. We achieved good results with traditional as well as deep learning models. We'll go over the details of LSTM modeling in the next section.\\\\
\textbf{Naïve Model}\\
As described in section 3.1, the scores for individual questions changes infrequently for an average patient over the course of two consecutive clinician visits (this is especially true since the scores are integers and don't allow partial progression from say 4 to 3). This suggests a Naive Model where the predicted score is the same as previous visit's score to serve as a baseline for all modeling approaches.\\\\
\textbf{Modeling Approaches}\\
In its given form, this is a multi-label multi-class classification problem. However, we can treat question\_type as a covariate and transform this into a multi-class classification (with 12x datapoints). We can also train a separate model for each of the 12 questions which would also transform it into a multi-class classification problem. In a similar vein, we can model the problem as a regression  and round the outputs to the nearest integer in [0, 4]. Regression modeling is what worked the best in our case and this was our approach henceforth.\\\\
% Another way to formulate the problem is by noticing that in the training data ~77\% of the next visit's score is the same as that of the previous visit and of the remaining 23\%, 14\% has next visit's score = previous visit's score - 1. We can train an MLP classifier to learn the unchanging scores case from the rest. This becomes a simple binary classification problem (it requires post-processing to re-map the same vs diff prediction to actual scores).\\\\
\textbf{The Overfitting problem}\\
One thing that became clear in very early stages of modeling was that overfitting was going to be the most prevalent issue which was evident across a number of models like linear regression, random forest, kNN, gradient boosted trees etc \cite{scikit-learn}. To this end we also included models geared specifically towards regularization like ElasticNet and Lasso \cite{hastie01statisticallearning}. The hyperparameter tuning was done via a grid search in the nested CV procedure described above.\\\\
\textbf{Training Data} \\
For the traditional ML models, the data was prepared to look as shown in Table 1, with all but last column as features and the last column as the target variable. Also note that this modeling was done on a per-question basis i.e. the \(previous\_value\) and \(future\_value\) in this table are for the question being modeled.\\
% ElasticNet and Lasso are both modified LinearRegression models with regularization term in their objective function in addition to the loss term. ElasticNet minimizes the following expression:\\
% \[1 / (2 * n\_samples) * ||y - Xw||^2_2 + l1\_weight * ||w||_1 + l2\_weight * ||w||^2_2\]
% where \(l1\_weight = alpha * l1\_ratio\) and \(l2\_weight = 0.5 * alpha * (1 - l1\_ratio)\)\\

% and its special case is Lasso when \(l1\_ratio=1.0\). An important parameter here is \(alpha\), which is the regularization strength. In ElasticNet \(l1\_ratio\) allows application of penalty on L1 as well as L2 norm of \(w\). 

% We use GridSearch \cite{scikit-learn} to tune hyperparameters for ElasticNet and Lasso models with \(alpha\) and \(l1\_ratio\) being tunable hyperparameters for ElasticNet and just \(alpha\) for Lasso.\\\\

\setlength\extrarowheight{5pt}
\begin{table}
\caption{Representative training data after pre-processing.}
\begin{tabular}{ | m{1.5cm} | m{2cm} | m{1.5cm} | m{1.5cm} | m{1.5cm} | m{0.5cm} | m{1.5cm} || m{1cm} |} 
  \hline
  \textbf{Patient ID} & \textbf{Days since Diagnosis} & \textbf{Previous Value} & \textbf{Delta days in Future} & \textbf{Sensor Feature1} &  \textbf{...} & \textbf{Sensor featureN} & \textbf{Future Value} \\
 \hline\hline
 \textbf{patient1} & 800 & 4 & 95 & 0.1 & ... & 22 & 3 \\ 
 \hline
 \textbf{patient2} & 700 & 3 & 90 & 0.4 & ... & 89 & 1 \\
 \hline
 \textbf{patient3} & 1400 & 2 & 120 & 0.2 & ... & 43 & 2 \\
 \hline
 \textbf{patient4} & 300 & 4 & 110 & 0.9 & ... & 09 & 4 \\
 \hline
 \textbf{patient5} & 500 & 4 & 100 & 0.0 & ... & 51 & 2 \\
  \hline
\end{tabular}

\end{table}

\subsubsection{LSTM}
\label{sec:lstm_modeling}

\begin{figure}%
    \centering
    \includegraphics[width=\linewidth]{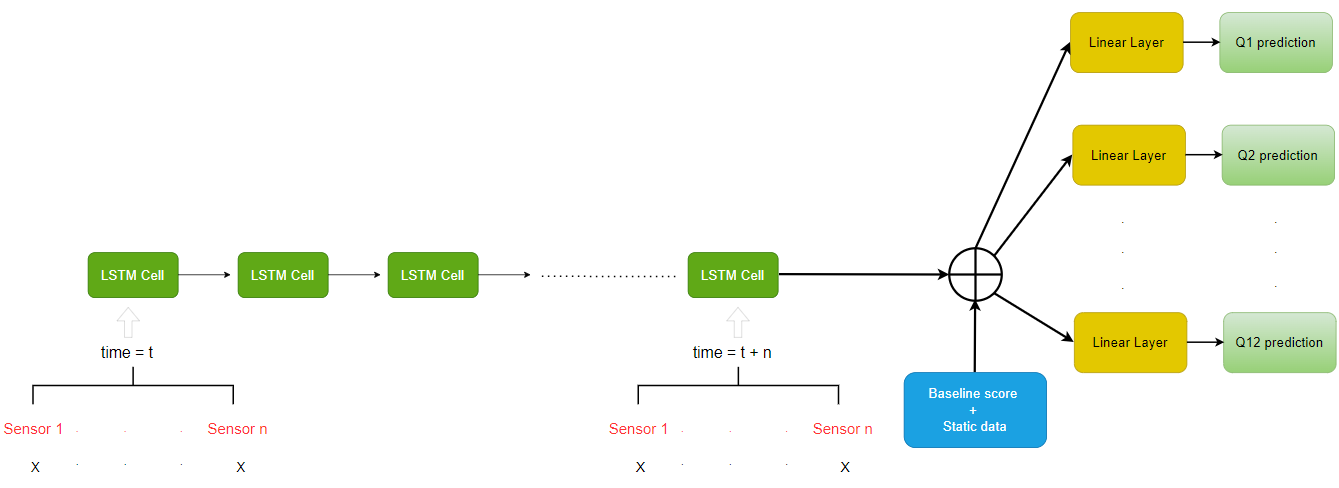}
    \caption{Architecture of the LSTM model used for predicting ALSFRS-R scores. The model processes time-series sensor data through sequential LSTM cells, capturing temporal dependencies. The output is combined with baseline scores and static patient data, then fed into multiple linear layers to predict each of the twelve ALSFRS-R sub-scores.}%
    \label{fig:LSTM}%
\end{figure}

In our study, we employed a Long Short-Term Memory (LSTM) neural network to predict ALSFRS-R scores based on time-series sensor data. 
The LSTM model is well-suited for handling sequential data due to its ability to capture long-term dependencies. 
As depicted in Figure \ref{fig:LSTM}, our model processes sensor data collected over multiple time points using a series of LSTM cells. 
Each cell captures temporal patterns in the data at different time steps, which are then fed into subsequent cells. 
To handle the uneven number of days between baseline and target scores, padding was applied to standardize the input sequences.
The output from the final LSTM cell is concatenated with baseline scores and static patient data to enhance the model's predictive capability. 
This combined feature set is then passed through multiple linear layers, each dedicated to predicting one of the twelve ALSFRS-R sub-scores. 
This architecture allows the model to leverage both dynamic sensor inputs and static information, providing robust predictions for each functional domain assessed by the ALSFRS-R score.

\section{Results}
\subsection{Task1}
The results for Task1 are shown in Table 2 below.

\setlength\extrarowheight{7pt}
\begin{table}
\caption{Task1's RMSE for various models separately modeling each of the 12 questions in ALSFRS-R scores. Green boxes denote the best performing model for the question. FS = FeatureSet, denoting whether median sensor features are used or TsFresh features are used (more details in section3.2. T1 denotes that only Task1 data was used for training whereas T1+2 denotes that Task1 data was augmented with that of Task2 for training. EN denotes ElasticNet model.}
\begin{tabular}{ | m{0.5cm} | m{2cm} | m{0.5cm} | m{0.5cm} | m{0.5cm} | m{0.5cm} | m{0.5cm} | m{0.5cm} | m{0.5cm} | m{0.5cm} | m{0.5cm} | m{0.5cm} | m{0.5cm} | m{0.5cm} |} 
\hline
 \textbf{FS} & \textbf{Model} & \textbf{Q1} & \textbf{Q2} & \textbf{Q3} & \textbf{Q4} & \textbf{Q5} & \textbf{Q6} & \textbf{Q7} & \textbf{Q8} & \textbf{Q9} & \textbf{Q10} & \textbf{Q11} & \textbf{Q12} \\
 \hline\hline
 \centering\textbf{-} & \textbf{Naive} & \cellcolor[HTML]{4fd313}0.47 & \cellcolor[HTML]{4fd313}0.38 & 0.43 & \cellcolor[HTML]{4fd313}0.65 & 0.69 & 0.77 & 0.74 & \cellcolor[HTML]{4fd313}0.62 & \cellcolor[HTML]{4fd313}0.78 & \cellcolor[HTML]{4fd313}0.5 & \cellcolor[HTML]{4fd313}0.46 & 0.54 \\
 \hline
 {\multirow{4}{*}{\rotatebox[origin=c]{90}{\textbf{W/o TsFresh}}}} & \textbf{T1 EN} & \cellcolor[HTML]{4fd313}0.47 & \cellcolor[HTML]{4fd313}0.38 & 0.43 & 0.67 & 0.73 & 0.75 & 0.77 & 0.62 & 0.85 & 0.59 & 0.47 & 0.65 \\
 \cline{2-14}
 & \textbf{T1 Lasso} & \cellcolor[HTML]{4fd313}0.47 & \cellcolor[HTML]{4fd313}0.38 & 0.43 & 0.66 & 0.73 & 0.79 & 0.79 & 0.64 & 0.79 & 0.55 & 0.49 & 0.54 \\
 \cline{2-14}
 & \textbf{T1+2 EN} & 0.54 & 0.57 & 0.51 & 0.79 & 0.93 & 0.74 & 0.74 & 0.66 & 0.85 & 0.66 & 0.66 & \cellcolor[HTML]{4fd313}0.42 \\
 \cline{2-14}
 & \textbf{T1+2 Lasso} & 0.54 & 0.56 & 0.49 & 0.66 & 0.85 & 0.7 & \cellcolor[HTML]{4fd313}0.69 & 0.64 & 0.86 & 0.71 & 0.66 & \cellcolor[HTML]{4fd313}0.42 \\
 \hline
 {\multirow{4}{*}{\rotatebox[origin=c]{90}{\textbf{W/ TsFresh}}}} & \textbf{T1 EN} & 0.49 & \cellcolor[HTML]{4fd313}0.38 & \cellcolor[HTML]{4fd313}0.42 & 0.69 & \cellcolor[HTML]{4fd313}0.69 & 0.79 & 0.78 & 0.67 & 0.82 & 0.5 & 0.48 & 0.54 \\
 \cline{2-14}
 & \textbf{T1 Lasso} & \cellcolor[HTML]{4fd313}0.47 & \cellcolor[HTML]{4fd313}0.38 & \cellcolor[HTML]{4fd313}0.42 & 0.71 & 0.69 & 0.8 & 0.77 & 0.67 & 0.81 & 0.52 & \cellcolor[HTML]{4fd313}0.46 & 0.46 \\
 \cline{2-14}
 & \textbf{T1+2 EN} & 0.54 & 0.54 & 0.52 & 0.69 & 0.95 & \cellcolor[HTML]{4fd313}0.69 & 0.72 & 0.72 & 0.98 & 0.69 & 0.58 & 0.76 \\
 \cline{2-14}
 & \textbf{T1+2 Lasso} & 0.57 & 0.55 & 0.52 & 0.67 & 0.8 & 0.69 & 0.73 & 0.69 & 0.88 & 0.7 & 0.64 & 0.72 \\
 \hline
\end{tabular}

\end{table}

The table displays the RMSE of various models on each question on test data split from the train+val set. As can be seen here as well as Figure 5, previous\_value has by far the most predictive power. For submitting, we ran Grid search using cross validation on entire training data (without the test split) and the model that gave the best validation set RMSE was selected for that question.

Using the above methodology, our final model was \textbf{ElasticNet + Naive} model. This achieved an RMSE of \textbf{0.5048} and MAE of \textbf{0.2222} on the final unseen test set and the leaderboard. This model was just shy of the Naive model submission that we made which had an RMSE of \textbf{0.4912} and MAE of \textbf{0.2024}.

Our LSTM model produced an RMSE of \textbf{0.8249} and MAE of \textbf{0.4761}, which, unfortunately, was not as strong as the naive model or the ElasticNet/Lasso.
This could be attributed to several factors, including the relatively small size of the dataset, which may have hindered the model's ability to capture complex temporal patterns. 
Additionally, the high variability in ALS progression among patients and potential noise in the sensor data could have impacted the model's performance.
\subsection{Task2}
Due to time constraints, we didn't manage to submit our model for Task2. However, using the ground truth that was released after the competition deadline, we ran our ElasticNet + Naive model and it gave an RMSE of \textbf{0.5594} and MAE of \textbf{0.2803}. This is better than the top leaderboard submission with RMSE of 0.5774 and MAE of 0.2879.
\begin{figure}
    \centering
    \subfloat[Task1 Q7]{{\includegraphics[width=4cm,height=4cm]{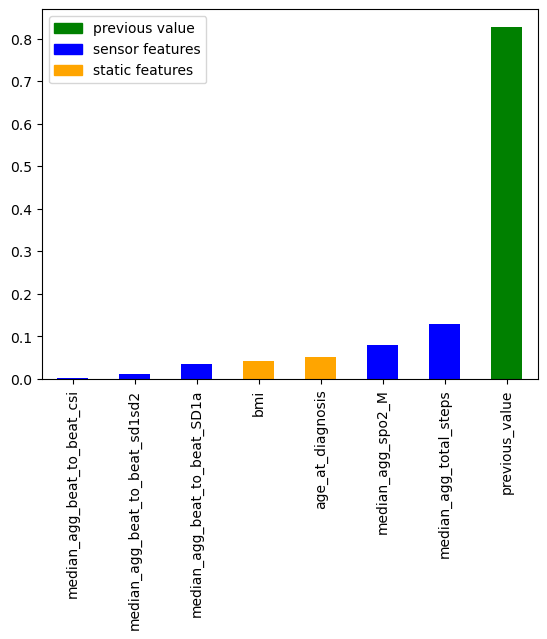} }}%
    \qquad
    \subfloat[Task1 TsFresh Q6]{{\includegraphics[width=4cm,height=4cm]{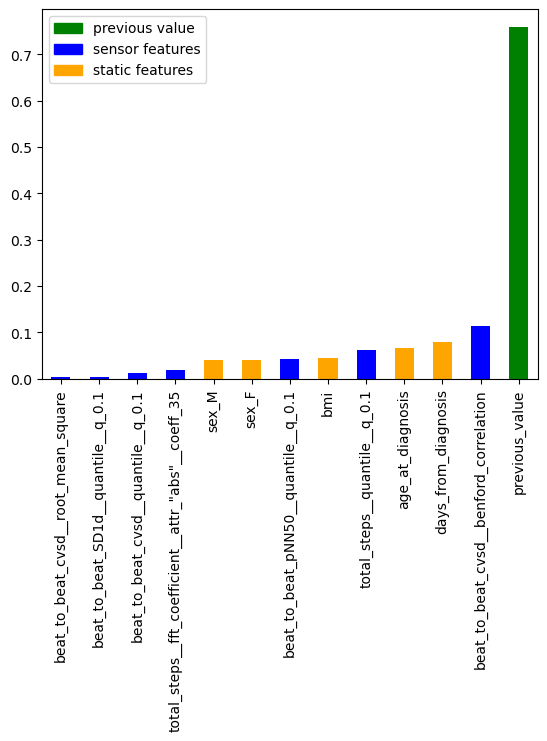} }}%
    \qquad
    \subfloat[Task2 Q5]{{\includegraphics[width=4cm,height=4cm]{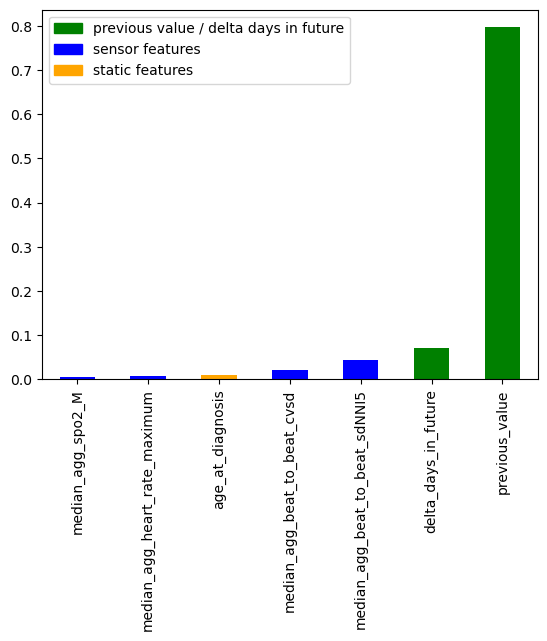} }}%
    \caption{Feature Importance charts showing "previous\_value" is consistently the most important feature.}%
    \label{fig:importance}%
\end{figure}
\subsection{Takeaways}
Following is a summary of our takeaways:
\begin{itemize}
    \item \textbf{Previous score} is an overwhelming contributor in prediction of the future score
    \item No single model could capture the essence of \textbf{all questions} better than the Naive model
    \item Small training set means that simpler models that focus on \textbf{regularization} perform well
    \item Having more granularity in scores either through having \textbf{floating point scores} or \textbf{larger span like [0, 10]} rather than [0, 4] would have allowed better modeling results, as 5 integer scale for scores seemed to mask inherent trends.
    \item Ablating \textbf{sensor data} in training didn't deteriorate the model performance much, which is evident from the Feature Importance scores graph in Figure 5
    \item \textbf{Notable features} that showed up in Feature Importances (other than previous\_value) across multiple questions include:
    \begin{itemize}
        \item \raggedright static features: age\_at\_diagnosis and FVC
        \item \raggedright median sensor features: agg\_respiration\_alpha2\_DimMean, agg\_respiration\_RMSSD, agg\_respiration\_SD1 and agg\_total\_steps
        \item \raggedright \texttt{tsfresh} sensor features: beat\_to\_beat\_cvsd\_\_benford\_correlation, beat\_to\_beat\_cvsd\_\_quantile\_\_q\_0.6 and total\_steps\_\_quantile\_\_q\_0.1
    \end{itemize}
\end{itemize}

% \section{Discussion}

% Discussion of the results and their implications.

\section{Future Work}
Future work will focus on clustering patient data to capture distinct patient phenotypes, allowing for more personalized predictions and treatment plans. 
We also plan to utilize freely available datasets such as PRO-ACT to enhance the robustness and generalizability of our models. 
Additionally, access to multimodal data sources such as audio data for speech, accelerometer data for muscle function, genetic data, imaging data, etc could help further improve the accuracy and reliability of ALSFRS-R score predictions. 
These efforts aim to refine our models and contribute to more effective ALS management and patient care.

Another point we plan to work on is integrating models for panel data. We've already experimented with this approach, where instead of fitting separate models for each question, we fitted a panel model on the whole data and defined the grouping variable as the question, the patient, or a nested combination of both.  The preliminary results from this approach were worse than the ``per-question modeling" approach. However, we believe that following a method similar to that of \cite{hajjem2014mixed} could still be beneficial. Such a methodology would make use of our existing models for the fixed effects part and augment them with random effects, e.g by treating the question as the grouping variable.  

\section{Conclusions}

We explored various machine learning approaches to predict ALSFRS-R scores. The main goal of our study was to determine if the app data provides additional value that enhances the predictive accuracy of the models, based on evidence found in the dataset that we analyzed.

The methodology involved extensive data preprocessing to synchronize clinical and sensor data, ensuring the integrity and reliability of the dataset. We generated features from both static and dynamic data sources, with a particular focus on leveraging temporal dependencies in the sensor data through techniques like median aggregation, feature-based time series analysis, and LSTM neural networks.

Our comparative analysis revealed that while on Task1 the naive baseline model achieved slightly better predictive accuracy, the ElasticNet+Naive regression model offered a robust framework for understanding feature contributions. Moreover, on Task2 our model (though not submitted) was able to perform slightly better than the Naive model. Nonetheless, the previous value of the ALSFRS-R score emerged as a significant predictor across all models.
Thus, our analysis did not find significant evidence that the app data provides an overall enhancement to the predictive accuracy of the model. Based on the leaderboard at the end of the competition, it would appear that the other teams reached a similar conclusion. 

We caution against discarding the potential usefulness of the app data at this stage however, as the dataset brought several methodological challenges with it, the biggest of which was its very small size.  We addressed it through rigorous cross-validation strategies and data augmentation techniques, ensuring that our models generalize well to unseen patients. The feature importance analysis highlighted and added value of certain static features and, for some questions, the importance of isolated sensor-derived features.

Overall, this work demonstrates the potential of integrating sensor data with machine learning models to enhance the monitoring and prediction of ALS progression, within the limits of a small dataset.

Given the strong persistence of the ``previous value" in both our modeling, and the lack of significant improvement to the baseline model, we hypothesise that the dataset would benefit from a larger and more heterogeneous set of patients, not only to increase the amount of observations, but to also introduce more variability in the target variable, perhaps spread across more regions and with clinical evaluations done by a multitude of clinicians. 

\section*{Acknowledgements}

The authors would like to extend their sincere gratitude to Dr. Jay Summet and the DS@GT CLEF team at Georgia Tech for their invaluable support and insights.

\bibliography{main}

\begin{thebibliography}{22}
\expandafter\ifx\csname natexlab\endcsname\relax\def\natexlab#1{#1}\fi
\providecommand{\url}[1]{\texttt{#1}}
\providecommand{\href}[2]{#2}
\providecommand{\path}[1]{#1}
\providecommand{\DOIprefix}{doi:}
\providecommand{\ArXivprefix}{arXiv:}
\providecommand{\URLprefix}{URL: }
\providecommand{\Pubmedprefix}{pmid:}
\providecommand{\doi}[1]{\href{http://dx.doi.org/#1}{\path{#1}}}
\providecommand{\Pubmed}[1]{\href{pmid:#1}{\path{#1}}}
\providecommand{\bibinfo}[2]{#2}
\ifx\xfnm\relax \def\xfnm[#1]{\unskip,\space#1}\fi
%Type = Inproceedings
\bibitem[{Birolo et~al.(2024{\natexlab{a}})Birolo, Bosoni, Faggioli, Aidos, Bergamaschi, Cavalla, Chiò, Dagliati, {de Carvalho}, {Di Nunzio}, Fariselli, {García Dominguez}, Marta~Gromicho, Longato, Madeira, Manera, Marchesin, Menotti, Silvello, Tavazzi, Tavazzi, Trescato, Vettoretti, Camillo, and Ferro}]{BiroloBosoniEtAl2024a}
\bibinfo{author}{G.~Birolo}, \bibinfo{author}{P.~Bosoni}, \bibinfo{author}{G.~Faggioli}, \bibinfo{author}{H.~Aidos}, \bibinfo{author}{R.~Bergamaschi}, \bibinfo{author}{P.~Cavalla}, \bibinfo{author}{A.~Chiò}, \bibinfo{author}{A.~Dagliati}, \bibinfo{author}{M.~{de Carvalho}}, \bibinfo{author}{G.~{Di Nunzio}}, \bibinfo{author}{P.~Fariselli}, \bibinfo{author}{J.~{García Dominguez}}, \bibinfo{author}{A.~G. Marta~Gromicho}, \bibinfo{author}{E.~Longato}, \bibinfo{author}{S.~Madeira}, \bibinfo{author}{U.~Manera}, \bibinfo{author}{S.~Marchesin}, \bibinfo{author}{L.~Menotti}, \bibinfo{author}{G.~Silvello}, \bibinfo{author}{E.~Tavazzi}, \bibinfo{author}{E.~Tavazzi}, \bibinfo{author}{I.~Trescato}, \bibinfo{author}{M.~Vettoretti}, \bibinfo{author}{B.~D. Camillo}, \bibinfo{author}{N.~Ferro},
\newblock \bibinfo{title}{{Overview of iDPP@CLEF 2024: The Intelligent Disease Progression Prediction Challenge}},
\newblock in: \bibinfo{booktitle}{Working Notes of the Conference and Labs of the Evaluation Forum ({CLEF} 2024), Grenoble, France, September 9th to 12th, 2024}, \bibinfo{year}{2024}{\natexlab{a}}.
%Type = Inproceedings
\bibitem[{Birolo et~al.(2024{\natexlab{b}})Birolo, Bosoni, Faggioli, Aidos, Bergamaschi, Cavalla, Chiò, Dagliati, {de Carvalho}, {Di Nunzio}, Fariselli, {García Dominguez}, Marta~Gromicho, Longato, Madeira, Manera, Marchesin, Menotti, Silvello, Tavazzi, Tavazzi, Trescato, Vettoretti, Camillo, and Ferro}]{BiroloBosoniEtAl2024b}
\bibinfo{author}{G.~Birolo}, \bibinfo{author}{P.~Bosoni}, \bibinfo{author}{G.~Faggioli}, \bibinfo{author}{H.~Aidos}, \bibinfo{author}{R.~Bergamaschi}, \bibinfo{author}{P.~Cavalla}, \bibinfo{author}{A.~Chiò}, \bibinfo{author}{A.~Dagliati}, \bibinfo{author}{M.~{de Carvalho}}, \bibinfo{author}{G.~{Di Nunzio}}, \bibinfo{author}{P.~Fariselli}, \bibinfo{author}{J.~{García Dominguez}}, \bibinfo{author}{A.~G. Marta~Gromicho}, \bibinfo{author}{E.~Longato}, \bibinfo{author}{S.~Madeira}, \bibinfo{author}{U.~Manera}, \bibinfo{author}{S.~Marchesin}, \bibinfo{author}{L.~Menotti}, \bibinfo{author}{G.~Silvello}, \bibinfo{author}{E.~Tavazzi}, \bibinfo{author}{E.~Tavazzi}, \bibinfo{author}{I.~Trescato}, \bibinfo{author}{M.~Vettoretti}, \bibinfo{author}{B.~D. Camillo}, \bibinfo{author}{N.~Ferro},
\newblock \bibinfo{title}{{Intelligent Disease Progression Prediction: Overview of iDPP@CLEF 2024}},
\newblock in: \bibinfo{booktitle}{Experimental {IR} Meets Multilinguality, Multimodality, and Interaction - 15th International Conference of the {CLEF} Association, {CLEF} 2024, Grenoble, France, September 9-12, 2024, Proceedings}, \bibinfo{year}{2024}{\natexlab{b}}.
%Type = Article
\bibitem[{Gupta et~al.(2023)Gupta, Patel, Premasiri, and Vieira}]{Gupta2023AthomeWA}
\bibinfo{author}{A.~S. Gupta}, \bibinfo{author}{S.~Patel}, \bibinfo{author}{A.~S. Premasiri}, \bibinfo{author}{F.~G. Vieira},
\newblock \bibinfo{title}{At-home wearables and machine learning sensitively capture disease progression in amyotrophic lateral sclerosis},
\newblock \bibinfo{journal}{Nature Communications} \bibinfo{volume}{14} (\bibinfo{year}{2023}). \URLprefix \url{https://api.semanticscholar.org/CorpusID:261062809}.
%Type = Article
\bibitem[{Vieira et~al.(2022)Vieira, Venugopalan, Premasiri, McNally, Jansen, McCloskey, Brenner, and Perrin}]{Vieira2022AMB}
\bibinfo{author}{F.~G. Vieira}, \bibinfo{author}{S.~Venugopalan}, \bibinfo{author}{A.~S. Premasiri}, \bibinfo{author}{M.~McNally}, \bibinfo{author}{A.~Jansen}, \bibinfo{author}{K.~McCloskey}, \bibinfo{author}{M.~P. Brenner}, \bibinfo{author}{S.~Perrin},
\newblock \bibinfo{title}{A machine-learning based objective measure for als disease severity},
\newblock \bibinfo{journal}{NPJ Digital Medicine} \bibinfo{volume}{5} (\bibinfo{year}{2022}). \URLprefix \url{https://api.semanticscholar.org/CorpusID:246240960}.
%Type = Article
\bibitem[{Pancotti et~al.(2022)Pancotti, Birolo, Rollo, Sanavia, Di~Camillo, Manera, Chi{\`o}, and Fariselli}]{pancotti2022deep}
\bibinfo{author}{C.~Pancotti}, \bibinfo{author}{G.~Birolo}, \bibinfo{author}{C.~Rollo}, \bibinfo{author}{T.~Sanavia}, \bibinfo{author}{B.~Di~Camillo}, \bibinfo{author}{U.~Manera}, \bibinfo{author}{A.~Chi{\`o}}, \bibinfo{author}{P.~Fariselli},
\newblock \bibinfo{title}{Deep learning methods to predict amyotrophic lateral sclerosis disease progression},
\newblock \bibinfo{journal}{Scientific reports} \bibinfo{volume}{12} (\bibinfo{year}{2022}) \bibinfo{pages}{13738}.
%Type = Article
\bibitem[{Choi et~al.(2015)Choi, Bahadori, Schuetz, Stewart, and Sun}]{Choi2015DoctorAP}
\bibinfo{author}{E.~Choi}, \bibinfo{author}{M.~T. Bahadori}, \bibinfo{author}{A.~Schuetz}, \bibinfo{author}{W.~F. Stewart}, \bibinfo{author}{J.~Sun},
\newblock \bibinfo{title}{Doctor ai: Predicting clinical events via recurrent neural networks},
\newblock \bibinfo{journal}{JMLR workshop and conference proceedings} \bibinfo{volume}{56} (\bibinfo{year}{2015}) \bibinfo{pages}{301--318}. \URLprefix \url{https://api.semanticscholar.org/CorpusID:5842463}.
%Type = Inproceedings
\bibitem[{Vidal et~al.(2023)Vidal, Sharpnack, Pinedo, Tsai, Lee, and Mart{\'\i}nez-L{\'o}pez}]{Vidal2023ComparativePA}
\bibinfo{author}{G.~Vidal}, \bibinfo{author}{J.~Sharpnack}, \bibinfo{author}{P.~Pinedo}, \bibinfo{author}{I.~C. Tsai}, \bibinfo{author}{A.~R. Lee}, \bibinfo{author}{B.~Mart{\'\i}nez-L{\'o}pez},
\newblock \bibinfo{title}{Comparative performance analysis of three machine learning algorithms applied to sensor data registered by a leg-attached accelerometer to predict metritis events in dairy cattle},
\newblock volume~\bibinfo{volume}{4}, \bibinfo{publisher}{Frontiers}, \bibinfo{year}{2023}, p. \bibinfo{pages}{1157090}.
%Type = Article
\bibitem[{Smedley et~al.(2018)Smedley, Ellingson, Cloughesy, and Hsu}]{Smedley2018LongitudinalPI}
\bibinfo{author}{N.~F. Smedley}, \bibinfo{author}{B.~M. Ellingson}, \bibinfo{author}{T.~F. Cloughesy}, \bibinfo{author}{W.~Hsu},
\newblock \bibinfo{title}{Longitudinal patterns in clinical and imaging measurements predict residual survival in glioblastoma patients},
\newblock \bibinfo{journal}{Scientific Reports} \bibinfo{volume}{8} (\bibinfo{year}{2018}). \URLprefix \url{https://api.semanticscholar.org/CorpusID:52846175}.
%Type = Book
\bibitem[{Pinheiro and Bates(2006)}]{pinheiro2006mixed}
\bibinfo{author}{J.~Pinheiro}, \bibinfo{author}{D.~Bates}, \bibinfo{title}{Mixed-effects models in S and S-PLUS}, \bibinfo{publisher}{Springer science \& business media}, \bibinfo{year}{2006}.
%Type = Article
\bibitem[{Hajjem et~al.(2014)Hajjem, Bellavance, and Larocque}]{hajjem2014mixed}
\bibinfo{author}{A.~Hajjem}, \bibinfo{author}{F.~Bellavance}, \bibinfo{author}{D.~Larocque},
\newblock \bibinfo{title}{Mixed-effects random forest for clustered data},
\newblock \bibinfo{journal}{Journal of Statistical Computation and Simulation} \bibinfo{volume}{84} (\bibinfo{year}{2014}) \bibinfo{pages}{1313--1328}.
%Type = Article
\bibitem[{Cascarano et~al.(2023)Cascarano, Mur-Petit, Hernandez-Gonzalez, Camacho, de~Toro~Eadie, Gkontra, Chadeau-Hyam, Vitria, and Lekadir}]{cascarano2023machine}
\bibinfo{author}{A.~Cascarano}, \bibinfo{author}{J.~Mur-Petit}, \bibinfo{author}{J.~Hernandez-Gonzalez}, \bibinfo{author}{M.~Camacho}, \bibinfo{author}{N.~de~Toro~Eadie}, \bibinfo{author}{P.~Gkontra}, \bibinfo{author}{M.~Chadeau-Hyam}, \bibinfo{author}{J.~Vitria}, \bibinfo{author}{K.~Lekadir},
\newblock \bibinfo{title}{Machine and deep learning for longitudinal biomedical data: a review of methods and applications},
\newblock \bibinfo{journal}{Artificial Intelligence Review} \bibinfo{volume}{56} (\bibinfo{year}{2023}) \bibinfo{pages}{1711--1771}.
%Type = Article
\bibitem[{Hu et~al.(2023)Hu, Wang, Drovandi, and Cao}]{hu2023predictions}
\bibinfo{author}{S.~Hu}, \bibinfo{author}{Y.-G. Wang}, \bibinfo{author}{C.~Drovandi}, \bibinfo{author}{T.~Cao},
\newblock \bibinfo{title}{Predictions of machine learning with mixed-effects in analyzing longitudinal data under model misspecification},
\newblock \bibinfo{journal}{Statistical Methods \& Applications} \bibinfo{volume}{32} (\bibinfo{year}{2023}) \bibinfo{pages}{681--711}.
%Type = Inproceedings
\bibitem[{W{\"o}rtwein et~al.(2023)W{\"o}rtwein, Allen, Sheeber, Auerbach, Cohn, and Morency}]{wortwein2023neural}
\bibinfo{author}{T.~W{\"o}rtwein}, \bibinfo{author}{N.~B. Allen}, \bibinfo{author}{L.~B. Sheeber}, \bibinfo{author}{R.~P. Auerbach}, \bibinfo{author}{J.~F. Cohn}, \bibinfo{author}{L.-P. Morency},
\newblock \bibinfo{title}{Neural mixed effects for nonlinear personalized predictions},
\newblock in: \bibinfo{booktitle}{Proceedings of the 25th International Conference on Multimodal Interaction}, \bibinfo{year}{2023}, pp. \bibinfo{pages}{445--454}.
%Type = Incollection
\bibitem[{Fulcher(2018)}]{fulcher2018feature}
\bibinfo{author}{B.~D. Fulcher},
\newblock \bibinfo{title}{Feature-based time-series analysis},
\newblock in: \bibinfo{booktitle}{Feature engineering for machine learning and data analytics}, \bibinfo{publisher}{CRC press}, \bibinfo{year}{2018}, pp. \bibinfo{pages}{87--116}.
%Type = Article
\bibitem[{Christ et~al.(2018)Christ, Braun, Neuffer, and Kempa-Liehr}]{christ2018time}
\bibinfo{author}{M.~Christ}, \bibinfo{author}{N.~Braun}, \bibinfo{author}{J.~Neuffer}, \bibinfo{author}{A.~W. Kempa-Liehr},
\newblock \bibinfo{title}{Time series feature extraction on basis of scalable hypothesis tests (tsfresh--a python package)},
\newblock \bibinfo{journal}{Neurocomputing} \bibinfo{volume}{307} (\bibinfo{year}{2018}) \bibinfo{pages}{72--77}.
%Type = Article
\bibitem[{Christ et~al.(2016)Christ, Kempa{-}Liehr, and Feindt}]{DBLP:journals/corr/ChristKF16}
\bibinfo{author}{M.~Christ}, \bibinfo{author}{A.~W. Kempa{-}Liehr}, \bibinfo{author}{M.~Feindt},
\newblock \bibinfo{title}{Distributed and parallel time series feature extraction for industrial big data applications},
\newblock \bibinfo{journal}{CoRR} \bibinfo{volume}{abs/1610.07717} (\bibinfo{year}{2016}). \URLprefix \url{http://arxiv.org/abs/1610.07717}. \href{http://arxiv.org/abs/1610.07717}{{\tt arXiv:1610.07717}}.
%Type = Article
\bibitem[{Pearson(1900)}]{Pearson1900}
\bibinfo{author}{K.~Pearson},
\newblock \bibinfo{title}{X. on the criterion that a given system of deviations from the probable in the case of a correlated system of variables is such that it can be reasonably supposed to have arisen from random sampling},
\newblock \bibinfo{journal}{The London, Edinburgh, and Dublin Philosophical Magazine and Journal of Science} \bibinfo{volume}{50} (\bibinfo{year}{1900}) \bibinfo{pages}{157--175}. \URLprefix \url{https://doi.org/10.1080/14786440009463897}. \DOIprefix\doi{10.1080/14786440009463897}.
%Type = Article
\bibitem[{Cawley and Talbot(2010)}]{cawley2010over}
\bibinfo{author}{G.~C. Cawley}, \bibinfo{author}{N.~L. Talbot},
\newblock \bibinfo{title}{On over-fitting in model selection and subsequent selection bias in performance evaluation},
\newblock \bibinfo{journal}{The Journal of Machine Learning Research} \bibinfo{volume}{11} (\bibinfo{year}{2010}) \bibinfo{pages}{2079--2107}.
%Type = Book
\bibitem[{Kuhn and Johnson(2019)}]{kuhn2019feature}
\bibinfo{author}{M.~Kuhn}, \bibinfo{author}{K.~Johnson}, \bibinfo{title}{Feature engineering and selection: A practical approach for predictive models}, \bibinfo{publisher}{Chapman and Hall/CRC}, \bibinfo{year}{2019}.
%Type = Misc
\bibitem[{Lyashenko and Jha(2024)}]{lyashenko_jha_2024}
\bibinfo{author}{V.~Lyashenko}, \bibinfo{author}{A.~Jha}, \bibinfo{title}{Cross-validation in machine learning: How to do it right}, \bibinfo{year}{2024}. \URLprefix \url{https://neptune.ai/blog/cross-validation-in-machine-learning-how-to-do-it-right}, \bibinfo{note}{accessed: 2024-05-24}.
%Type = Article
\bibitem[{Pedregosa et~al.(2011)Pedregosa, Varoquaux, Gramfort, Michel, Thirion, Grisel, Blondel, Prettenhofer, Weiss, Dubourg, Vanderplas, Passos, Cournapeau, Brucher, Perrot, and Duchesnay}]{scikit-learn}
\bibinfo{author}{F.~Pedregosa}, \bibinfo{author}{G.~Varoquaux}, \bibinfo{author}{A.~Gramfort}, \bibinfo{author}{V.~Michel}, \bibinfo{author}{B.~Thirion}, \bibinfo{author}{O.~Grisel}, \bibinfo{author}{M.~Blondel}, \bibinfo{author}{P.~Prettenhofer}, \bibinfo{author}{R.~Weiss}, \bibinfo{author}{V.~Dubourg}, \bibinfo{author}{J.~Vanderplas}, \bibinfo{author}{A.~Passos}, \bibinfo{author}{D.~Cournapeau}, \bibinfo{author}{M.~Brucher}, \bibinfo{author}{M.~Perrot}, \bibinfo{author}{E.~Duchesnay},
\newblock \bibinfo{title}{Scikit-learn: Machine learning in {P}ython},
\newblock \bibinfo{journal}{Journal of Machine Learning Research} \bibinfo{volume}{12} (\bibinfo{year}{2011}) \bibinfo{pages}{2825--2830}.
%Type = Book
\bibitem[{Hastie et~al.(2001)Hastie, Tibshirani, and Friedman}]{hastie01statisticallearning}
\bibinfo{author}{T.~Hastie}, \bibinfo{author}{R.~Tibshirani}, \bibinfo{author}{J.~Friedman}, \bibinfo{title}{The Elements of Statistical Learning}, Springer Series in Statistics, \bibinfo{publisher}{Springer New York Inc.}, \bibinfo{address}{New York, NY, USA}, \bibinfo{year}{2001}.

\end{thebibliography}

% \appendix
% \section{Online Resources}

\end{document}